%% file: main.tex
\documentclass[runningheads]{llncs}

\usepackage{eccv}

\usepackage{eccvabbrv}

\usepackage{graphicx}
\usepackage{booktabs}
\usepackage{wrapfig}

\usepackage[accsupp]{axessibility}  

\usepackage{hyperref}

\usepackage{orcidlink}

\begin{document}

\title{ShapeFusion: A 3D diffusion model for localized shape editing} 


\author{Rolandos Alexandros Potamias\orcidlink{0000-0003-0049-2589} \and
Michail Tarasiou\orcidlink{0000-0002-6282-4529} \and
Stylianos Ploumpis\orcidlink{0000-0002-4836-1513}
\and Stefanos Zafeiriou\orcidlink{0000-0002-5222-1740}}

\authorrunning{R.A.~Potamias et al.}

\institute{Imperial College London, United Kingdom,\\
\email{\{r.potamias, michail.tarasiou10,s.ploumpis,s.zafeiriou\}@imperial.ac.uk}}

\maketitle

\begin{abstract}
  In the realm of 3D computer vision, parametric models have emerged as a ground-breaking methodology for the creation of realistic and expressive 3D avatars. Traditionally, they rely on Principal Component Analysis (PCA), given its ability to decompose data to an orthonormal space that maximally captures shape variations. However, due to the orthogonality constraints and the global nature of PCA's decomposition, these models struggle to perform localized and disentangled editing of 3D shapes, which severely affects their use in applications requiring fine control such as face sculpting. In this paper, we leverage diffusion models to enable diverse and fully localized edits on 3D meshes, while completely preserving the un-edited regions. We propose an effective diffusion masking training strategy that, by design, facilitates localized manipulation of any shape region, without being limited to predefined regions or to sparse sets of predefined control vertices. Following our framework, a user can explicitly set their manipulation region of choice and define an arbitrary set of vertices as handles to edit a 3D mesh. Compared to the current state-of-the-art our method leads to more interpretable shape manipulations than methods relying on latent code state, greater localization and generation diversity while offering faster inference than optimization based approaches. Project
page: \url{https://rolpotamias.github.io/Shapefusion/}.
\end{abstract}

\section{Introduction}
\label{sec:intro}

3D human bodies and faces are considered the core of a wide range of applications in the context of gaming, graphics and virtual reality in our modern digital avatar era \cite{zheng2023ilsh}. Over the last decade several methods have been developed to model 3D humans with parametric models being among the best performing ones \cite{egger20203d}. 
Parametric and 3D morphable models (3DMMs) attempt to project the 3D shapes in compact low-dimensional latent representation, usually via PCA, that is able to efficiently capture the essential characteristics and variations of the human shape \cite{booth20163d,SMPL,MANO,Handy,ploumpis2020towards}. 
Recently, several methods have shown that non-linear models such as Graph Neural Networks \cite{ranjan2018generating,Spiral} or implicit functions \cite{giebenhain2023nphm} can further improve the modeling of 3D shapes. However, fundamentally all the aforementioned methods share a common limitation; an entangled latent space that hinders localized editing. Using an entangled latent space, parametric models are not human interpretable and therefore it remains non-trivial to identify latent codes that are able to control region specific features. 

\begin{figure*}[!t]
    \centering 
    \includegraphics[width=\textwidth]{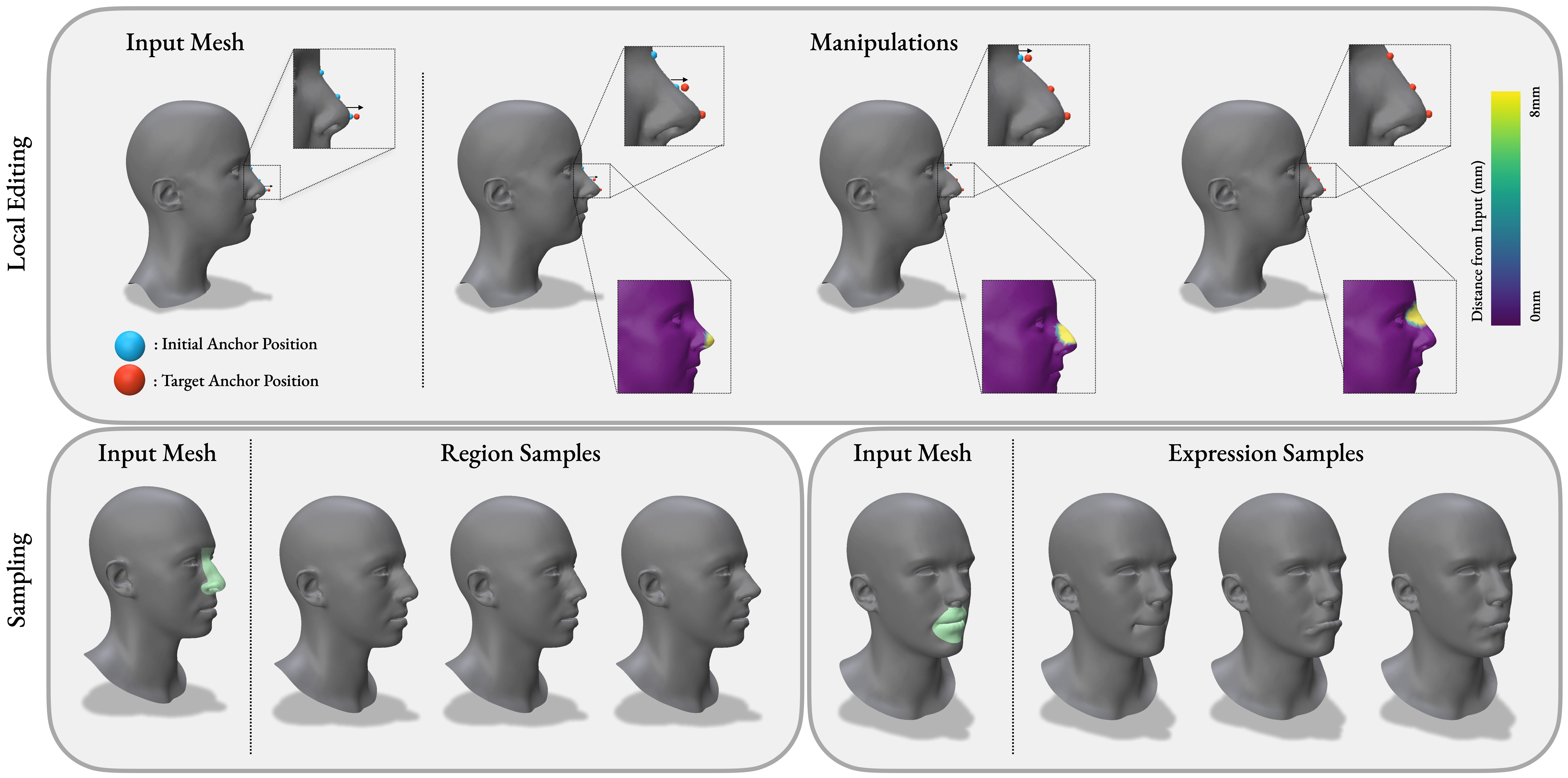}
    \captionof{figure}{Illustration of the properties of the proposed method for localized editing (Top) and region sampling (Bottom).  Top:  The proposed method can manipulate any region of a mesh by simply setting a user-defined anchor point and its surrounding region. The manipulations are completely disentangled and affect only the selected region. The disentanglement of the manipulations is illustrated using the color-coded distances from the previous manipulation step. Bottom: The proposed method can also sample new face parts and expressions by simply defining a mask over the desired region.}
    \label{fig:examples_face}
\end{figure*}

The global nature of current state-of-the-art parametric models poses a major limitation in the process of realistic human avatar generation that requires localized manipulation. To enable editing and region specific manipulation, experienced 3D artists are required to further process and sculpt the generated avatar. Recently, several methods have attempted to generalize  neural parametric models for local editing by enforcing disentanglement in the factorized latent space \cite{foti2022sd_vae,foti2023led}. However, albeit making a step towards localized editing, both models attempt to solve the disentanglement task in the latent space, which apart from causing a significant reconstruction performance drop, can not guarantee disentanglement in the 3D space. Furthermore, local manipulation of shapes in the latent space, limits their interpretability thus constraining their practical use in real-world applications.
Following the success of diffusion models \cite{luo2021diffusion,zeng2022lion}, we propose a simple but effective technique for localized 3D human modeling, that extends prior works on point cloud diffusion models to 3D meshes, using a geometry-aware embedding layer. We formulate the task of localized shape modelling as an inpainting problem using a masking training approach, where the diffusion process acts locally on the masked regions. Using this masking strategy we enable learning of local topological features that facilitate manipulation directly on the vertex space and guarantees the disentanglement of the masked from the unmasked regions. Additionally, the proposed masking approach enables conditioned local editing and sculpting, simply by selecting an arbitrary set of anchor points to drive the generation process to the desired manipulations. This contrasts with present state-of-the-art models, that not only struggle to achieve effective localized 3D attribute manipulation but also necessitate a costly optimization process for direct control from sparse anchor points. 

\noindent To sum up the contributions of this study
can be summarized as: 
\begin{itemize}
    \item We introduce a simple but effective training strategy for diffusion models that learns local priors of the underlying data distribution 
    which highlights the superiority of diffusion models compared to traditional VAE architectures for localized shape manipulations. 
    
    \item We introduce a localized 3D model which enables direct point manipulation, sculpting and expression editing directly on the 3D space. ShapeFusion not only guarantees fully localized editing of the user-selected regions but also provides an interpretable paradigm compared to current methods which rely on the state of the latent code for mesh manipulation. ShapeFusion provides artists with a powerful neural-based 3D editing tool, enabling precise modifications to any area of a 3D shape by leveraging learned data priors.
    
    \item We showcase that ShapeFusion not only generates diverse region samples that outperform the current state-of-the-art models by a large margin, but also learns strong priors that can substitute current parametric models. 
\end{itemize}

\section{Related Work}
\label{sec:related}
\textbf{Disentangled and Localized Models.} 
Disentangled generative models aim to encode the underlying factors of variation in the data in disjoint subsets of features, allowing interpretable and independent control over each factor. Training disentangled models has been a long studied research field in computer vision \cite{ma2018disentangled,burgess2018understanding,kim2018disentangling}. Initial approaches used bi-linear models to learn disentangled representations of content and style \cite{tenenbaum2000separating}.  InfoGAN \cite{chen2016infogan} pioneered the disentangled modeling that enabled fine-grained control on the process. In \cite{mathieu2016disentangling}, the authors proposed an adversarial training pipeline to learn separable representations of labeled images. In the 3D domain, disentangled representations usually rely on separating shape and pose using supervised \cite{jiang2020disentangled} and unsupervised methods \cite{zhou2020unsupervised,chen2021intrinsic,lombardi2021latenthuman,mu2021sdf,hui2022neural} that rely on customized loss functions to enforce disentanglement. In \cite{sun2023learning}, the authors proposed a latent swapping loss to enforce the disentanglement of human joints and shape. Similarly, several methods have attempted to learn local facial expression manipulations using sparse and region-based PCA to animate static \cite{tena2011interactive} and dynamic faces \cite{neumann2013sparse,wu2016anatomically}. Recently, Qin \etal \cite{NFR} proposed a neural based rigging method that disentangles facial expressions from subject's identity to enable in-the-wild re-targeting. Although untangling shape and pose has been a long-standing area of study, achieving spatially disentangled shape editing remains a challenging task. Foti \etal \cite{foti2022sd_vae} attempted to disentangle face and body parts using a mini-batch swapping of the shape attributes, while enforcing consistency in the factorized latent space. In a follow up work \cite{foti2023led}, the authors proposed a local eigenprojection loss that enforces the orthogonality between latent variables which improved disentanglement performance. Recently, Tarasiou \etal \cite{tarasiou2024locally} attempted to tackle disentangled face editing using a sparse set of control points to guide the manipulation. However, all of the aforementioned works attempt to learn disentangled representations by factorizing the latent space which apart from reducing the model's reconstruction performance, it can not guarantee localized manipulation in the 3D shape. In contrast, we propose a method that tackles, by design, localized manipulation directly in the 3D space which compared to prior works, is fully interpretable and can guarantee spatially localized edits by its design. 

\noindent\textbf{Human Parametric Models.}
Parametric models are generative models that enable the generation of new shapes by modifying their compact latent representations, so called parameters. 
The first parametric model, was proposed by Blanz and Vetter \cite{blanz2023morphable}, pioneered the era of 3D morphable models by creating a face model from 200 scans. 
The authors proposed a global shape model that utilized principal component analysis (PCA) to encode the variations of the dataset. PCA has been shown to accurately fit the data distribution and enable the generation of new shape combinations. Following this work, several methods have been proposed to advance face modeling by using large scale datasets \cite{booth20163d,li2017learning} that better capture the population variations and additional head part scans \cite{hu2017avatar,dai20173d,ploumpis2019combining,ploumpis2020towards} that enable full head modeling. The success of PCA models to represent 3D shapes has also been established for modelling other parts of the human body, including the entire body \cite{anguelov2005scape,SMPL,STAR} and the hands \cite{MANO,Handy}. However, PCA models usually require a large amount of parameters to accurately model diverse datasets. To overcome such limitations, Ranjan \etal \cite{ranjan2018generating} proposed to learn human face variation using a compact neural network, with 75\% less parameters compared to PCA models. Recently, a neural implicit representation of human heads was proposed in \cite{giebenhain2023nphm}, that models human faces using continuous representations. The authors proposed to use a separate module to learn local regions of the face and combine the local features using a global aggregation module. However, similar to the aforementioned approaches the models learn global shape variations, that restrict localized editing. 

\noindent\textbf{Diffusion Models for 3D shape generation.}
Over the last years, diffusion models have revolutionized the field of image generation, providing powerful and flexible generative models, overcoming the limitations of Generative Adversarial Networks and Variational Autoencoders \cite{ho2020denoising,dhariwal2021diffusion,LDM}. Luo \etal \cite{luo2021diffusion} proposed the first diffusion model applied to 3D shapes by using the diffusion process to learn a conditional distribution of point clouds. The authors proposed to gradually insert noise in the point space and trained a denoising network to predict the inserted noise. In contrast, LION \cite{zeng2022lion} proposed to embed the input point cloud in two separate latent spaces that encode coarse and detailed shape features and train two distinct diffusion models on those latent vectors. In order to extend LION to mesh generation tasks, the authors utilized an off-the-shelf triangulation method \cite{SAP}. Recently, MeshDiffusion \cite{liu2023meshdiffusion} achieved remarkable results in 3D shape generation by leveraging the deformable tetrahedral grid parametrization \cite{shen2021dmtet}. However, such parametrization requires an initial time-consuming iterative fitting process which limits the applicability of the method. In this work, we extend \cite{luo2021diffusion} from point clouds to triangular meshes with fixed topology and enforce localized attribute learning using an inpainting technique during training. 

\section{Method}
\label{sec:method}
\begin{figure*}[!ht]
    \centering
    \includegraphics[width=\textwidth]{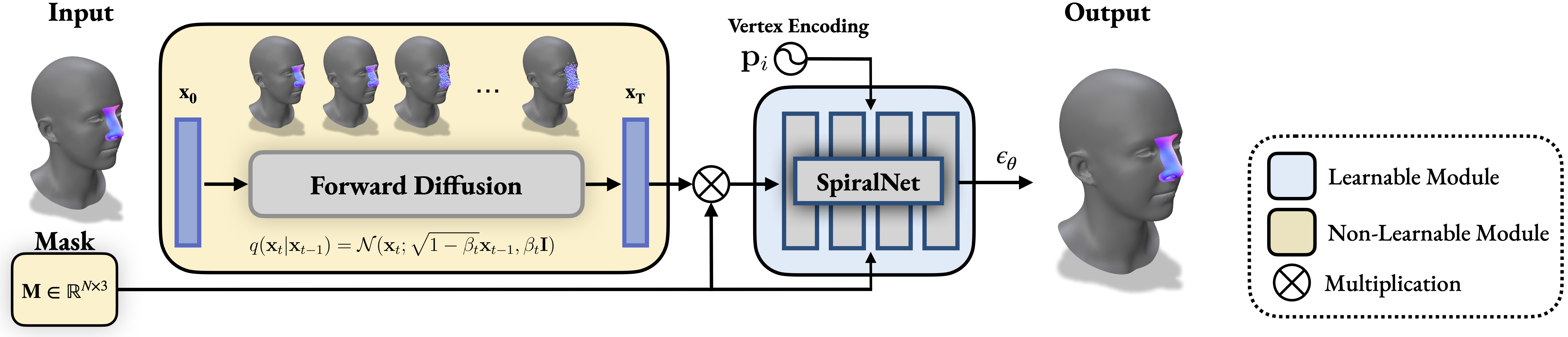}
    \captionof{figure}{
    \textbf{Method overview:} We propose a 3D diffusion model for localized attribute manipulation and editing. During forward diffusion step, noise is gradually added to random regions of the mesh, indicated by a mask $\mathbf{M}$. In the denoising step, a hierarchical network based on mesh convolution is used to learn a prior distribution of each attribute directly on the vertex space. 
    \label{fig:method}}
\end{figure*}

Motivated from the shortcomings of prior works to achieve fully localized 3D shape manipulation, we propose a training scheme that follows a masked diffusion process and construct a fully localized model that is able to guarantee local manipulations on the 3D space.
The proposed framework is composed of two main components: the Forward Diffusion process that gradually introduces noise to the input mesh and the Denoising Module that predicts the denoised version of the input. \cref{fig:method}
illustrates an overview of the proposed method.

\subsection{Forward Diffusion}
Having a mesh $\mathcal{M} = (\mathcal{V}, \mathcal{E})$ with $N$ vertices $\mathbf{x} \in \mathcal{V}$ and $E$ edges $\mathcal{E}$ defined from the faces of the mesh, the Forward Diffusion process gradually adds noise sampled from a Gaussian distribution $\mathcal{N}(\mu, \sigma\mathbf{I})$ to the input vertices. This process is repeated $T$ times as a Markov chain until the vertices are transformed into a Gaussian distribution $\mathcal{N}(\textbf{0}, \mathbf{I})$. Similar to \cite{ho2020denoising}, we define the forward diffusion process as: 
\begin{equation}
    q(\mathbf{x}_t | \mathbf{x}_{t-1}) =  \mathcal{N}(\mathbf{x}_t | \sqrt{1-\beta_t} \mathbf{x}_{t-1}, \beta_t\mathbf{I}), \quad t \in [1,T]
\end{equation}
where $\beta_t$ is the variance schedule parameter that controls the noise scheduling of the process.

In order to train a localized model we define a masked forward diffusion process that gradually adds noise to specific areas of the mesh as defined by a mask $\mathbf{M} \in \mathbb{R}^{N \times 3}$. 
During training, we define the masked vertices $\mathbf{M}$ as the $k$-hop geodesic neighborhood of a randomly selected \emph{anchor point} $\mathbf{x}_a \in \mathcal{V}$. The remaining vertices, including the anchor point, remain unaffected. 
Using this masked diffusion process we  guarantee local editing as well as full control of the generative process without employing an explicit conditional model. In contrast to the previous methods for disentangled manipulation, our approach not only guarantees fully localized editing but also enables direct manipulation of any point and region of the mesh.

\subsection{Denoising Module using Mesh Convolutions}
In the second stage of our method we train a denoising module $\epsilon_\theta$, that acts directly on the 3D space, to predict the noise $\epsilon_t$ added to the input, by employing the reparametrization of  \cite{ho2020denoising}: 
\begin{equation}
    \mathcal{L}_{t} = || \epsilon_t - \epsilon_\theta(\mathbf{x}, t, \mathbf{M})||_2 
\end{equation}
where $\epsilon_t$ is the noise at time-step $t$ of the forward diffusion process and $\mathbf{M}$ is a binary mask that defines the manipulated region. Unlike unstructured 3D point cloud generation, generating meshes from noise remains non-trivial. This is predominantly due to the structured nature of 3D meshes that requires careful design of the denoising module to respect the input topology.
\begin{wrapfigure}{r}{0.4\textwidth}
  \begin{center}
    \includegraphics[width=0.4\textwidth]{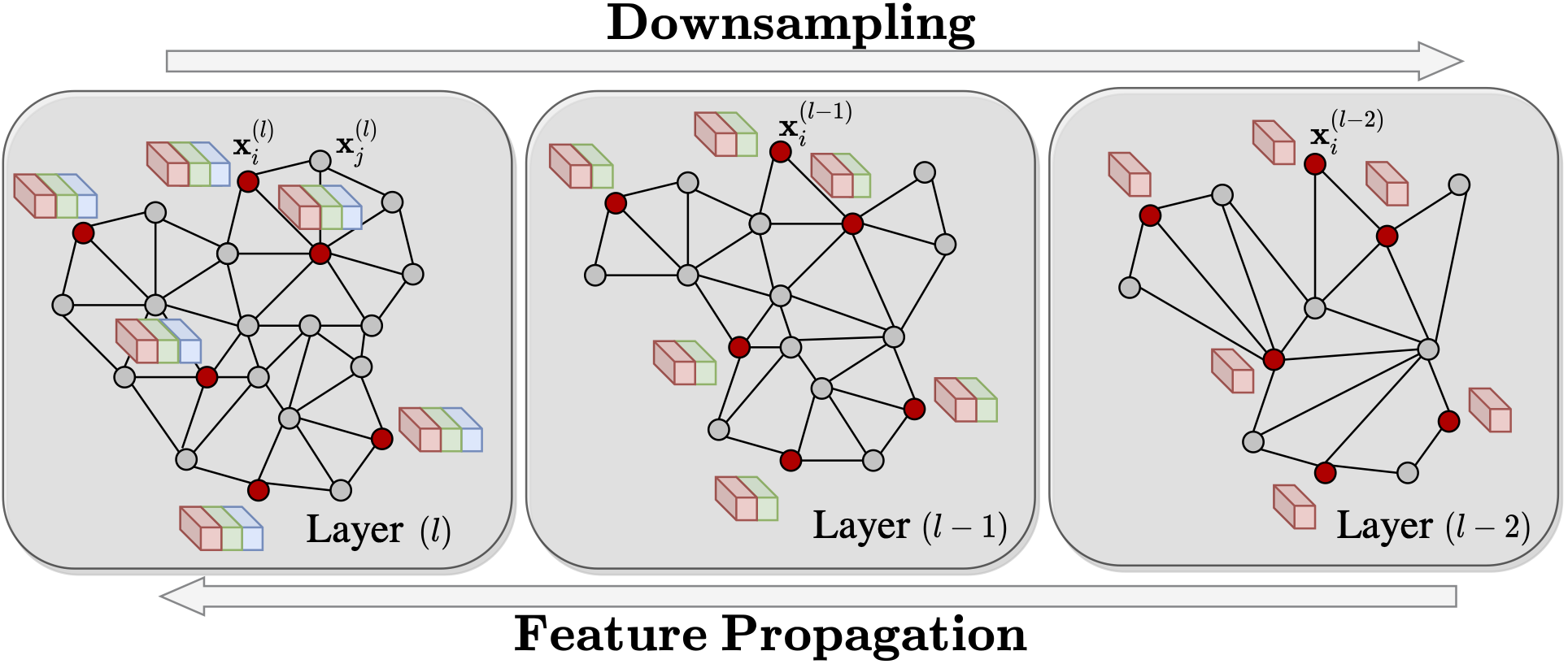}
  \end{center}
  \vspace{-0.1cm}
  \captionof{figure}{The proposed hierarchical message passing layer. At each layer the features are aggregated recursively from the coarser to the finer levels. Using such masking approach we can guarantee localized edits from the design of the method.}
  \label{fig:hierarchical}
\end{wrapfigure}
 In particular, using a simple permutation equivariant module, similar to point cloud diffusion methods \cite{luo2021diffusion,zeng2022lion}, would result in generation of unordered points that follow the input distribution without preserving mesh topology. Intuitively, this is attributed to the fact that the network cannot distinguish the noise of one point from another resulting in irregular triangulation.  To enforce topology preservation, for each vertex we introduce a vertex-index positional encoding $\mathbf{p}_i$ that explicitly defines the index number $i$ of the vertex on the mesh. In doing so, each vertex is paired with a positional encoding, indicating its index. This not only breaks the permutation equivariance of each layer but also enables the network to learn a vertex-specific prior  \cite{baltatzis2023neural}. 
Furthermore, we introduce a hierarchical mesh convolution layer to accomplish two main goals: a) allow information propagation between distant regions of the shape and b) enforce the manipulated regions to respect the unmasked geometry. The first is necessary for 3D shapes such as faces and bodies that have highly symmetrical regions whereas, the latter is a highly desirable property to ensure smoothness of the generated meshes. 
To allow long range dependencies between nodes on the mesh, for every layer of the proposed denoising model we utilize three hierarchical levels $({l})$ of mesh convolutions acting on different mesh resolutions. We recursively calculate the features of each node $\mathbf{x}_i$ from its features at different levels of resolution. Given that the network operates on a fixed mesh topology, the downsampled representations of the mesh can be pre-computed similar to \cite{ranjan2018generating},  and the hierarchical features can be efficiently calculated without any computational overhead. In this setting, we derive a nested hierarchical graph where the vertices at each coarser level $(l-1)$ constitute a distinct subset of the vertices at the finer level $(l)$, i.e. $\mathbf{x}^{(l)} \subset \mathbf{x}^{(l-1)}$. 
As shown in \cref{fig:hierarchical}, at each layer the message passing steps start from the coarser levels and recursively update the finer levels. The vertex $i$ features $\mathbf{f}_i^{(l)}$ at level $(l)$ can be then defined as: 
\begin{equation}
\label{eq:hierarchical}
    \mathbf{f}_i^{(l)} = \gamma \left( \sum_{j \in \mathcal{N}_i} \mathbf{W}^{(l)}_j \left(\sum_{k}^{(l-1)}  \mathbf{f}_j^{(k)} - \mathbf{f}_i^{(k)} \right) + \mathbf{b}^{(l)}_j \right)
\end{equation}
where $\mathbf{W}^{(l)}_j, \mathbf{b}^{(l)}_j$ are the learnable parameters of the level $(l)$ related to node $j$ in the neighborhood $\mathcal{N}_i$ of node $i$ and $\gamma$ is a non-linear activation function. The features of  vertex $i$, $\mathbf{f}_i$, are initialized on the first layer of the network as: 
\begin{equation}
    \mathbf{f}^{(0)}_i = \left[ \mathbf{x}_i || \mathbf{m}_i ||\mathbf{p}_i || \mathbf{c}_t \right]
\end{equation}
where $\mathbf{x}_i \in \mathbb{R}^{N\times3}$ are the $xyz$-coordinates of the vertex $i$, $\mathbf{p}_i$ the vertex-index positional encoding of the $i$-th vertex, $\mathbf{m}_i$ the input binary mask and $\mathbf{c}_t$ an embedding that corresponds to the $t$-timestep of the diffusion process. Similar to \cite{potamias2022graphwalks}, we utilized relative features to enhance the expressivity of the model. 
For a fair comparison, we follow \cite{foti2022sd_vae,foti2023led} and utilize spiral mesh convolution \cite{Spiral,potamias2020learning}, although any other graph neural network formulation could also be utilized. Using spiral mesh convolutions, the neighborhood of vertex $i$, $\mathcal{N}_i$, is now uniquely defined by the spiral ordering of the nodes around it. 

\section{Experiments}
\textbf{Datasets.} 
To train the proposed method we utilized three datasets that include human faces and bodies. Similar to Foti \etal \cite{foti2022sd_vae}, we used the publicly available \textsc{UHM} \cite{ploumpis2020towards} and \textsc{STAR} \cite{STAR} models and sampled 10K distinct identities of faces and bodies respectively. For each dataset, a 3D artist partitioned the template meshes into several regions corresponding to different body and face parts. Furthermore, to evaluate our model on registered 3D facial scans, we also utilized the MimicMe \cite{papaioannou2022mimicme} dataset which remains the largest publicly available 3D facial dataset consisting of approximately 5K subjects with diverse morphological characteristics. We followed a 90/10\% training/testing split for all the datasets. 

\noindent\textbf{Implementation Details.} For all experiments we utilize 4 layers of hierarchical mesh convolutions, each consisting of 3 resolution levels. All of the layers have 64-hidden units followed by ReLU activations. We implement the vertex-index positional encoding as a learnable embedding layer with 16-hidden units and Fourier positional embeddings to encode the diffusion timestep $t$ following \cite{luo2021diffusion}. The model is trained for 600 epochs using Adam optimizer \cite{kingma2014adam} with a linear learning rate decay schedule starting from $1e-2$.

\noindent\textbf{Baselines.} We compared the proposed method with SD  \cite{foti2022sd_vae} and LED \cite{foti2023led}, the current state-of-the-art models for disentangled manipulation. Additionally, we implemented a strong baseline method that uses a similar training formulation with the proposed method. In particular, we train a VAE method (M-VAE) based on Spiral Convolutions, with an architecture identical to \cite{foti2022sd_vae,foti2023led}, using the masking technique proposed in \cref{sec:method}. Under this setting the M-VAE takes as input a masked mesh and attempts to reconstruct the original mesh, which enforces the model to learn local features. 
\subsection{Localized Region Sampling}
\begin{figure}[!t]
    \centering 
    \includegraphics[width=0.95\linewidth]{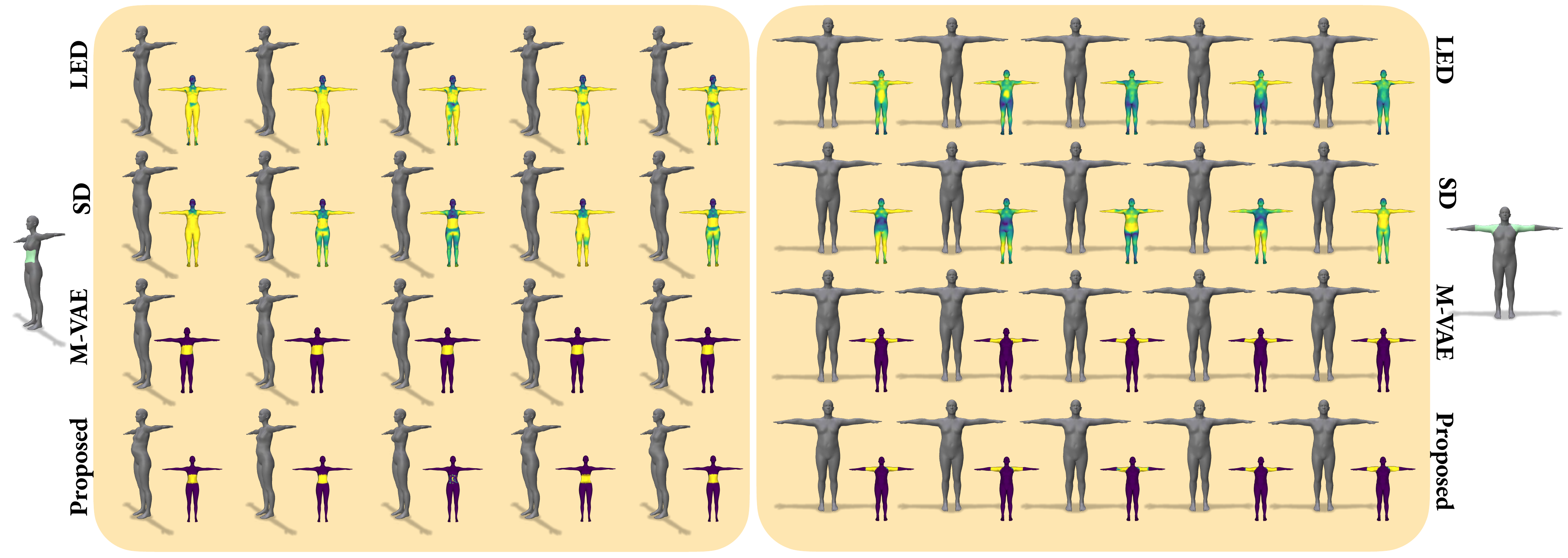}
    \captionof{figure}{\small{Qualitative and quantitative comparison between the proposed and the baseline methods on the STAR dataset. On the left sides we show the input meshes, with the manipulated region highlighted green. The region samples along with their heatmap are illustrated row-wise. Figure better viewed in zoom.}
    \label{fig:examples_star}}
\end{figure}

In this section we assess the generated outputs from both the proposed and the baseline models perform in terms of Diversity (DIV), Identity Preservation (ID) and Fréchet inception distance (FID). In particular, to measure the diversity of the sampled regions for each subject we sampled 10 different parts for 5 random manipulation regions and measured their mean square error from the original region. In this context, models with small diversity will only be able to generate regions that mimic the input. To evaluate the realism of the manipulated regions we implemented an FID \cite{heusel2017gans} inspired loss that calculates the Fréchet distance between the PCA projections of the generated and the ground truth meshes. Specifically, we trained a PCA 3DMM, similar to \cite{ploumpis2019combining,SMPL}, and projected the ground truth meshes, along with a set of random attribute manipulations for each ground truth mesh to the PCA latent space and measured the distance between the distributions of the real and the manipulated latent codes. In a similar manner, we evaluated the Identity Preservation by measuring the average per-subject MSE between the original and the manipulated PCA projections. For additional details about the utilized metrics we refer the reader to the supplementary material. 
In Table \ref{tab:results} we report the results of the proposed and the baseline methods on the MimicMe \cite{papaioannou2022mimicme}, UHM \cite{ploumpis2019combining} and STAR \cite{STAR} datasets. As can be seen, the proposed method outperforms the baseline methods under all metrics, even by a large margin. The proposed method achieves highly diverse manipulations that respect the boundary conditions while at the same time the editing remains realistic. In contrast, LED and SD methods fail to accurately preserve the subjects' identity and generate realistic manipulations of each region. These findings are further validated in \cref{fig:examples_face},  where we visualize 5 random samples generated from the proposed and the baseline methods, for different manipulated regions along with a color-coded distance map that indicates the distance from  the original mesh. 
A similar behaviour can also be observed in \cref{fig:examples_star}, where local manipulations on the arms and the belly affect distant regions such as the legs and the head. This is congruent with our hypothesis that segmenting the latent space results in poor reconstruction performance. Additionally, although the M-VAE produces better results in terms of FID and ID compared to the baseline methods, it fails to match the diversity of the proposed method. It is crucial to highlight that manipulations on the posed space are unnecessary since they can be achieved through a trivial canonicalization step that will map the articulated body to the canonical pose. 

\input{Tables/main_results}
As can be easily seen from the color-coded distance maps, on both \cref{fig:examples_face} and \cref{fig:examples_star}, the proposed method is the only one that achieves guaranteed localized manipulation without affecting any other region. Additionally, the generated samples of the proposed method exhibit large variations from the ground truth mesh that validates the generative power of the model. This is in contrast to M-VAE method that although it enables highly localized manipulations, the generated regions are almost identical. This can be attributed to the autoencoding structure of M-VAE, which mainly focuses on reconstructing the input, resulting in small deviations from the ground truth reconstruction. This is in line with the findings of Table \ref{tab:results}.
\begin{figure*}[!h]
    \centering 
    \includegraphics[width=0.95\textwidth]%
    {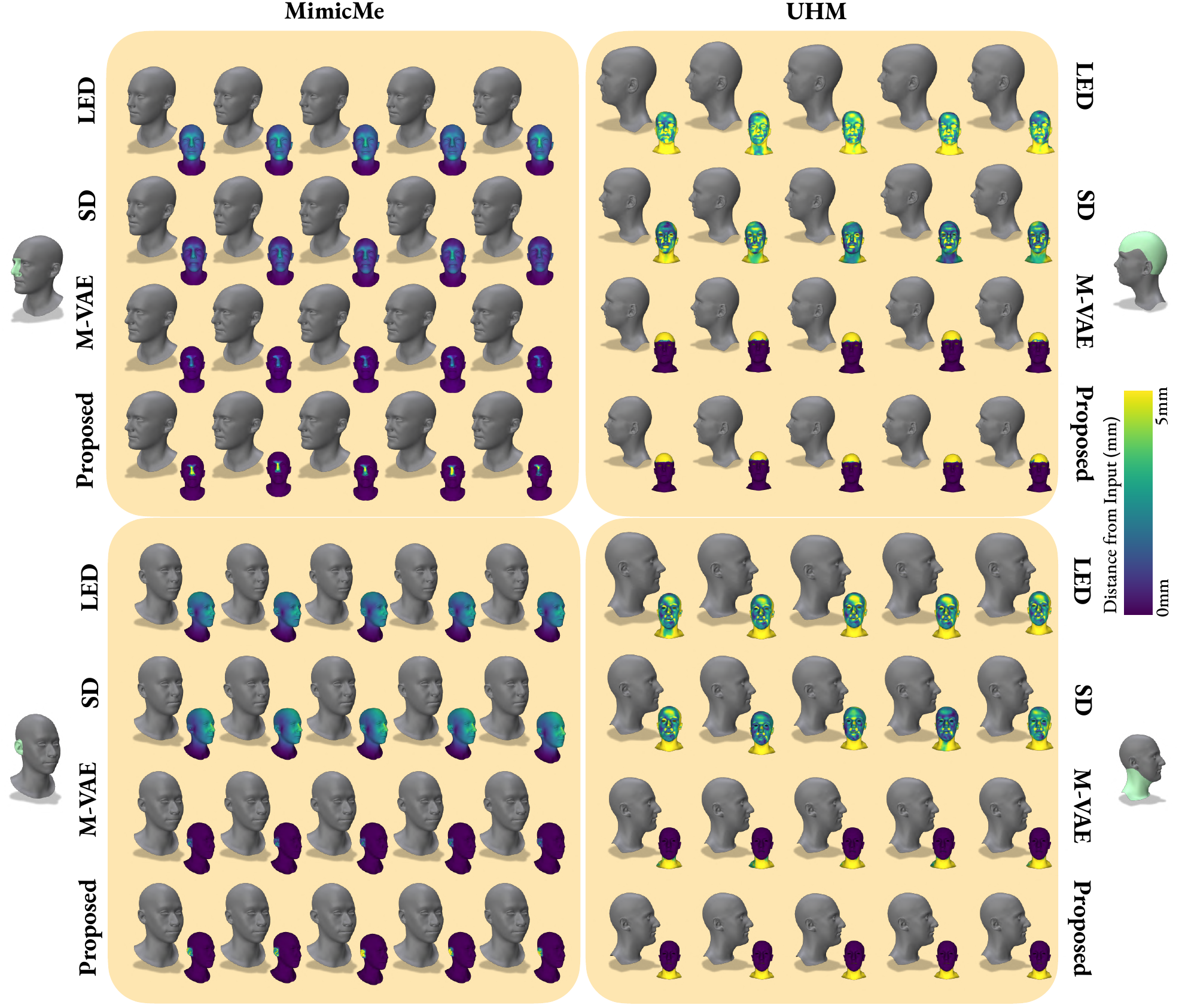}
    \captionof{figure}{\small{Qualitative and quantitative comparison between the proposed and the baseline methods. On the left and right sides we show the input meshes from MimicMe and UHM dataset respectively, with the manipulated region highlighted in green. In each of the rows we illustrate 5 samples generated from each method for the same region along with a heatmap indicating the differences with the original input. Please note that the proposed method achieves bigger displacements, which translates to more diverse samples, localized only on the manipulated region. Figure better viewed in zoom. For additional region manipulations we refer the reader to the supplementary material.}
    \label{fig:examples_face}}
\end{figure*}

\subsection{Direct point manipulation}
\begin{figure}[!h]
    \centering
    \includegraphics[width=0.9\linewidth]{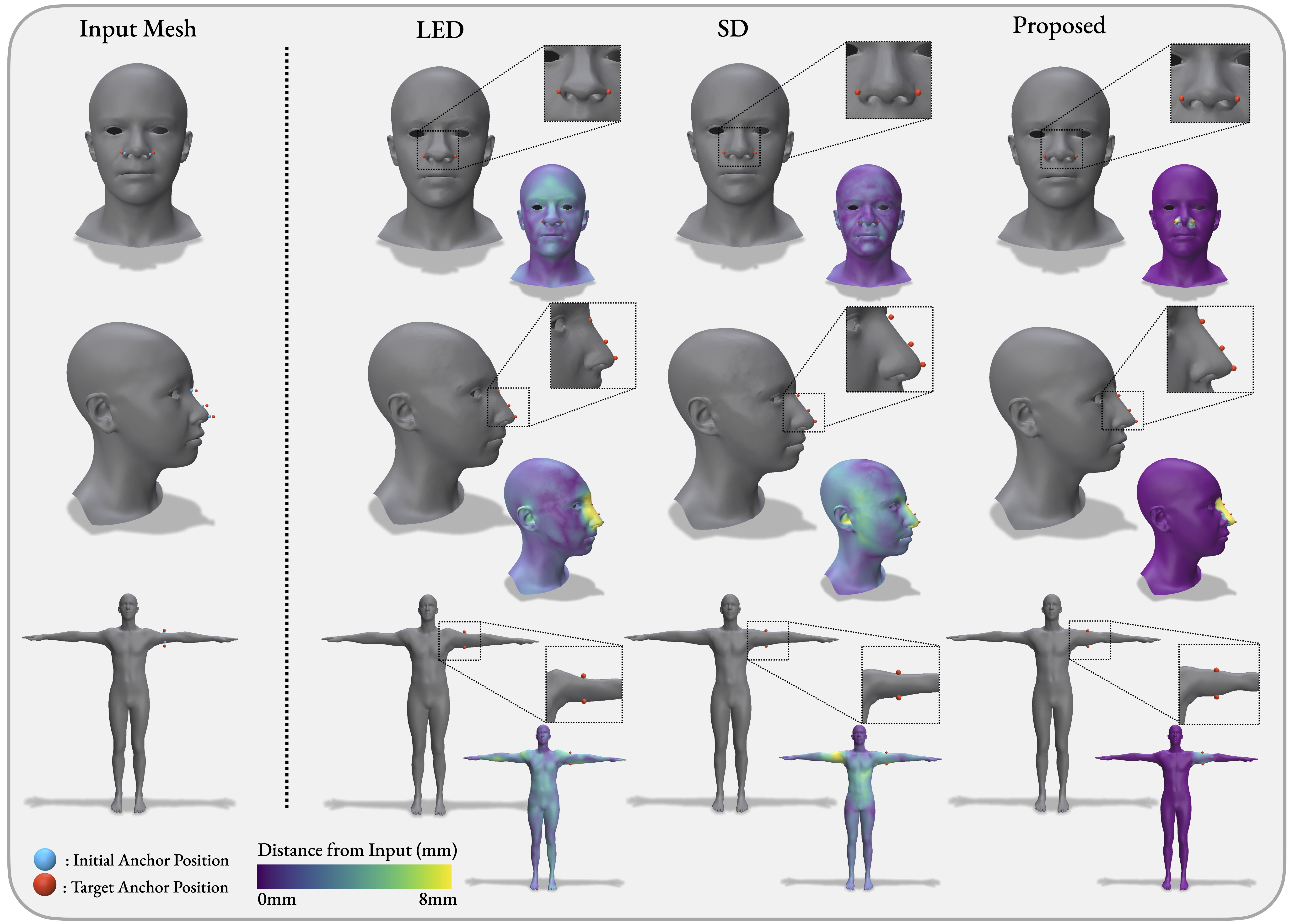}
    \captionof{figure}{\small{Left: Local editing of an input mesh from a set of anchor points (blue) and desired positions (red). Right: The generated manipulations of each method are displayed along with the desired anchor points positions (red) and a heatmap indicating the per-vertex distance with the input mesh. The proposed method inherently attains, by definition, zero error in the desired positions without requiring any optimization procedure and simultaneously achieves complete localization.}
    \label{fig:manipulation}}
\end{figure}

A pivotal property of the proposed method is its ability to locally edit any region of the mesh conditioned on a single point. This characteristic empowers the model to perform direct manipulations of any region by simply sliding an anchor point $\mathbf{x}_a$. Hence, a user can choose single or multiple vertices, define their desired new positions and feed them to the proposed method to generate a locally deformed mesh that follows the desired locations. In contrast,  previous methods \cite{foti2022sd_vae,foti2023led} required an optimization procedure to find the latent codes that minimize the distance from the target positions. The proposed method does not rely on any optimization procedure and can directly generate an edited mesh by setting the desired vertex positions and defining a mask $\textbf{M}$ which includes their surrounding region. In \cref{fig:manipulation} we compare the direct point manipulation of the proposed and the baseline methods. The proposed method attains fully localized editing by modifying only the region surrounding the anchor points defined by the mask. Both SD \cite{foti2022sd_vae} and LED \cite{foti2023led} methods exhibit limited disentanglement capabilities, as quantified on the heatmaps of \cref{fig:manipulation}. Additionally, it is worth noting that the proposed method can manipulate meshes approximately 10-times faster ($\sim3.2$sec) compared to the baseline methods ($\sim22$sec) that require a time-consuming optimization fitting process.  
\subsection{Global Sampling and Reconstruction}
In this section we evaluate the properties of the proposed model as a powerful prior for shape generation and reconstruction. Specifically, apart from localized region manipulation the proposed method can be utilized as a generative model for unconditioned face and body generation. 
Considering that the model was trained using randomly selected regions of varying size, the proposed method can be applied to effectuate the direct generation of complete shapes by masking the entire shape region. 
\begin{figure}[!ht]
    \centering
    \includegraphics[width=\linewidth]{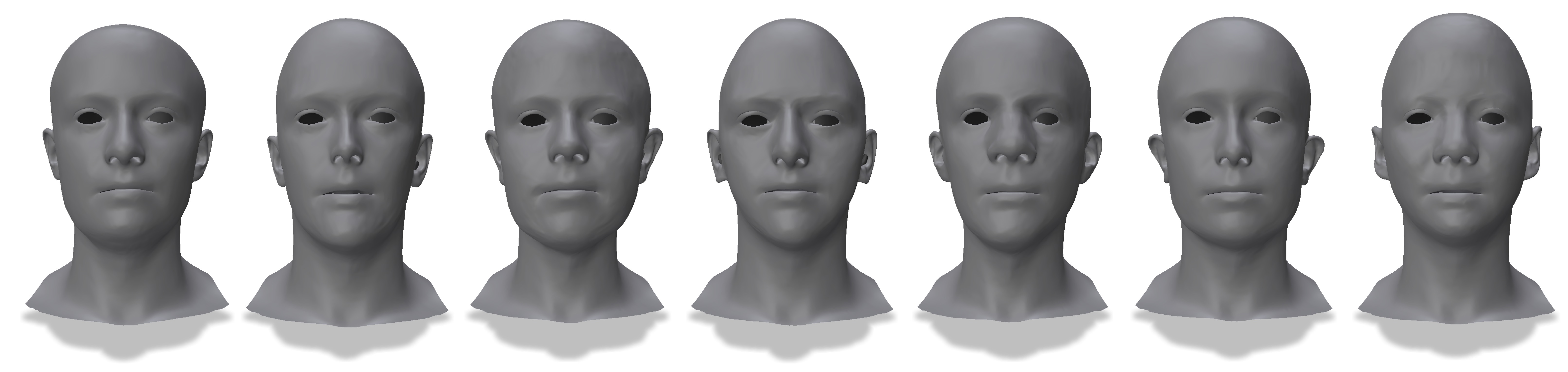}
    \vspace{-0.2cm}
    \captionof{figure}{Arbitrary samples generated from the proposed method.
    \label{fig:samples_from_noise}}
\end{figure}
As shown in \cref{fig:samples_from_noise}, the proposed model can produce a wide range of facial variations. For additional qualitative and quantitative results on global sampling performance, we refer the reader to the supplementary material. 
\begin{figure*}[!h]
    \centering
    \includegraphics[width=\linewidth]{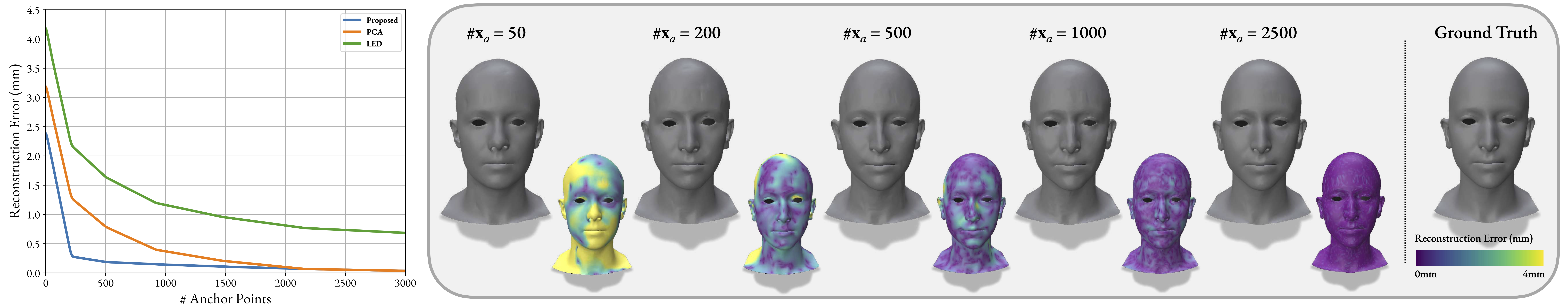}
    \vspace{-0.2cm}
    \captionof{figure}{Left: Quantitative evaluation of the proposed method on the UHM test set under a different number of anchor points \#$\mathbf{x}_a$. Right: Qualitative illustration of the effect of number of anchor points $\mathbf{x}_a$ to condition the reconstruction. With around 200 points the proposed method can reconstruct the details of the ground truth shape. 
    \label{fig:reconstruction}}
\end{figure*}
Furthermore, we quantitatively evaluated the proposed method as an autodecoder model to reconstruct a sparse input. In contrast to the autoencoder structure of most popular 3D shape models, the proposed method can reconstruct an input in an autodecoder setup by conditioning the generation process on sparse anchor points $\mathbf{x}_a$. To quantify the reconstruction performance of the proposed method, for each subject on the test set we sampled a variable set of anchor points and masked the rest of the shape, which was then reconstructed using our method. For comparison purposes, we conducted an optimization step to fit SD and PCA models to the set of anchor points. From \cref{fig:reconstruction} (Left) we observe that with 200 points the proposed method can provide an adequate representation of the input face, achieving 0.38mm reconstruction error, outperforming PCA and SD methods. This can also be validated in  \cref{fig:reconstruction} (Right) where the reconstruction of a random test sample is illustrated for a different number of anchor points. Using as few as 200 anchor points, the proposed method can effectively restore the facial shape identity, whereas utilizing a greater number of anchor points facilitates the representation of finer facial details.

\subsection{Region Swapping}
\begin{figure}[!h]
    \centering
    \includegraphics[width=\linewidth]{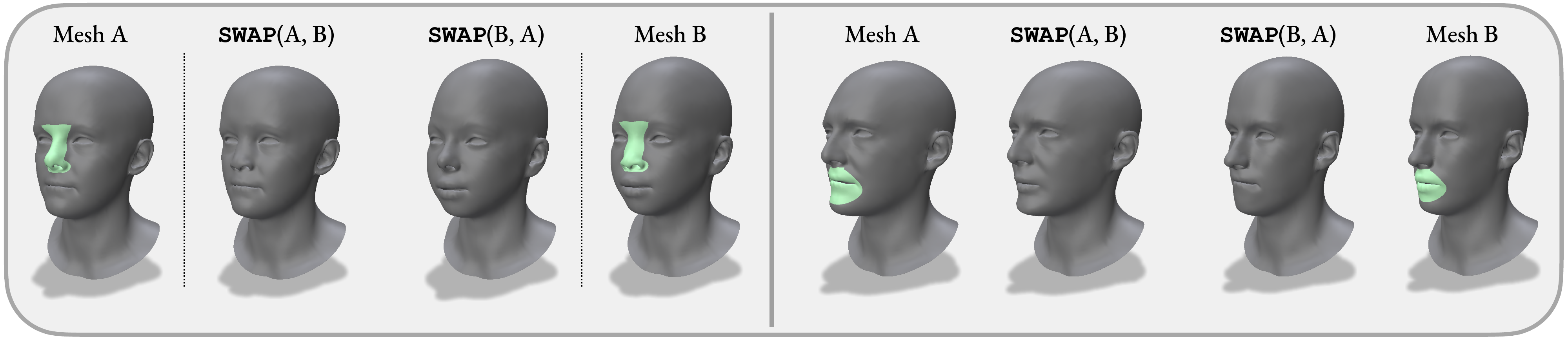}
    \captionof{figure}{Swaping facial regions between four random identities. We report swaping between the facial regions, highlighted in green, of Mesh A (left) to Mesh B (right) (\texttt{SWAP}(A,B)) and the opposite (\texttt{SWAP}(B,A)).  
    \label{fig:swap}}
\end{figure}

A practical property of the proposed method, with potential real-world applications in aesthetic medicine, is its ability to seamlessly swap distinct facial regions and components between different identities. Specifically, for a given source region on mesh A and a target region on mesh B, we condition the generation of the masked region on a set of anchor points defined from the target mesh B. To enable smooth swapping between the two face parts we avoid selecting anchor points on the boundaries of each region. In \cref{fig:swap}, we illustrate samples of region swapping from mesh A to mesh B, as well as the reverse operation.
\subsection{Localized Expression Manipulation}
\begin{figure}[!h]
    \centering
    \includegraphics[width=0.9\linewidth]{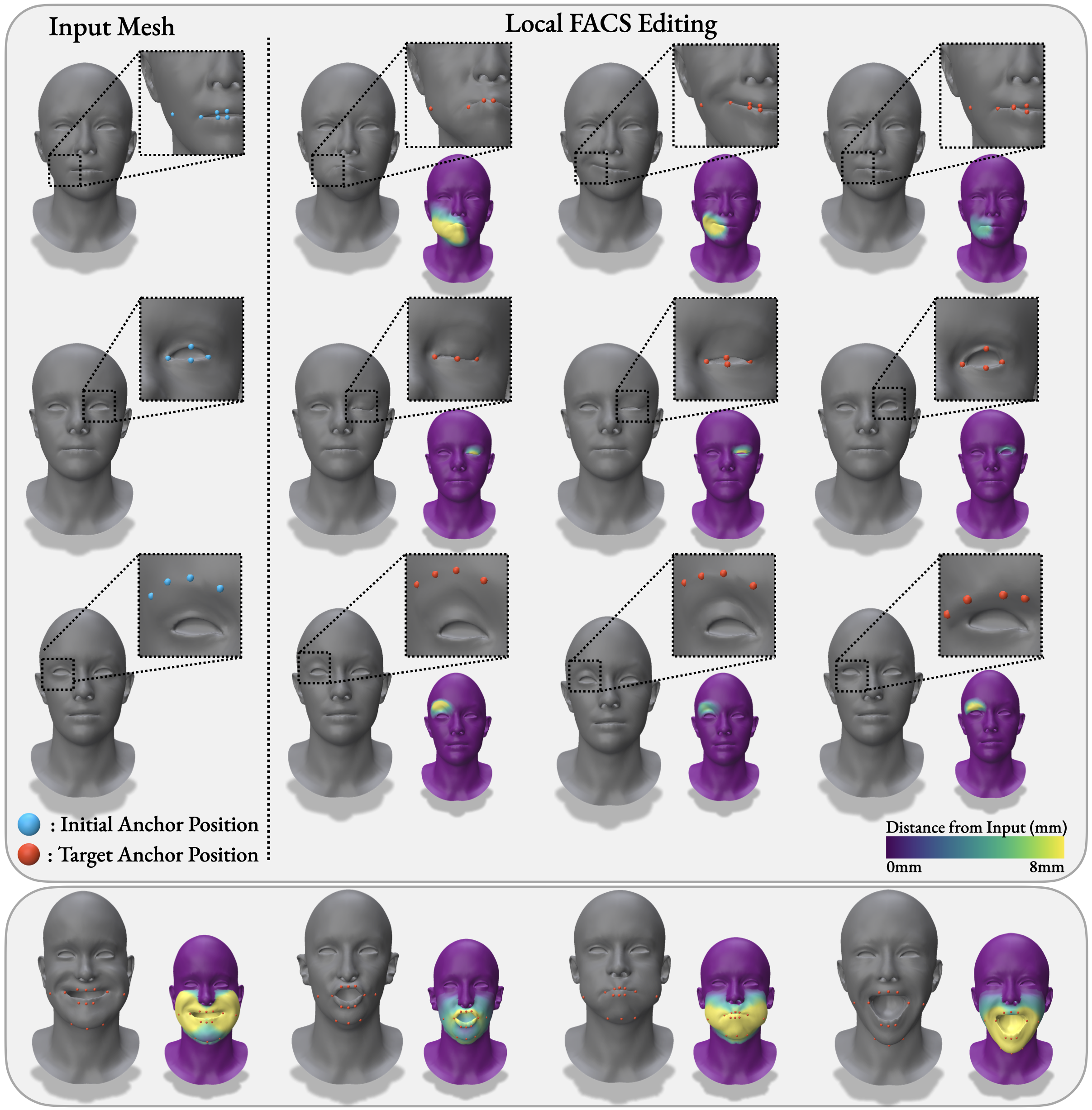}
    \captionof{figure}{Localized Expression Editing from a set of selected anchor points (blue) and desired positions (red). We showcase both small and extremely localized FACS manipulations (top) and larger regions that results to more extreme expressions (bottom). 
    \label{fig:expressions}}
\end{figure}

In addition to localized editing within the identity space, there is a notable gap in the existing literature concerning localized expression manipulation. Similar to parametric models, expression blendshapes rely on global PCA models, that restricts their ability for local and localized manipulation. Currently, localized expression editing necessitates the involvement of graphic artists to rig facial models based on anatomical muscle activity, known as the Facial Action Coding System (FACS) \cite{rosenberg2020face}. However, aside from the considerable manual effort required, FACS relies on specific pre-defined action units (AU), thereby imposing limitations on their local editing flexibility. 
In contrast, by employing the proposed diffusion model the benefit is two-folded. First, we can manipulate extreme expressions similar to existing expression editing methods \cite{NFR} as shown in the \cref{fig:expressions} (bottom). Secondly, we can also attain spatially localized manipulations of any point on the face, by simply adjusting the masked region, which remains a notable limitation of current manipulation methods.
To demonstrate that, we trained the proposed model on the expressions of MimicMe dataset \cite{papaioannou2022mimicme} using the same settings reported in \cref{sec:method}. \cref{fig:expressions} depicts the localized manipulations of a neutral face towards a target expression, as defined by the red anchor points. The proposed model can achieve fully-localized edits that affect only the user-defined manipulation region, as can be observed from the color-coded meshes. In particular, as shown in \cref{fig:expressions}, the proposed method can perform {spatially local} manipulation and generalize to out-of-distribution expressions, such as the smirk, that were not present in the training data.

\begin{figure}[!ht]
    \centering
    \includegraphics[width=\linewidth] {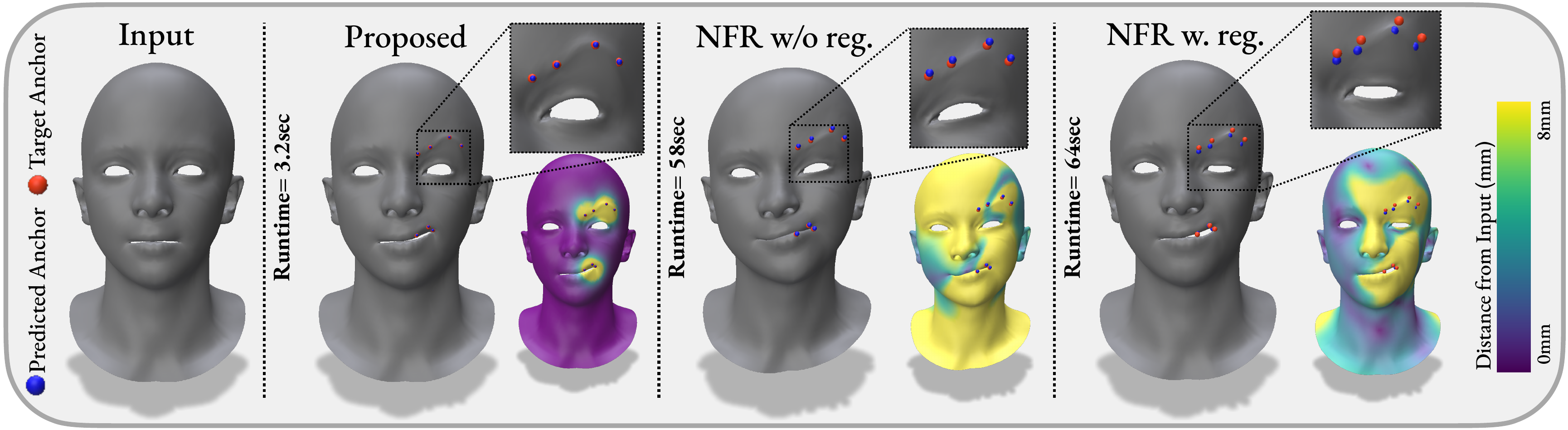}
    \captionof{figure}{Quantitative comparison between the proposed and the NFR \cite{NFR} methods on  expression editing given a set of desired positions (red). 
    \label{fig:nfr}}
\end{figure}

To further evaluate the expression manipulations of the proposed method, we compared it against NFR \cite{NFR}, the current state-of-the-art method for localized expression editing. In \cref{fig:nfr} we illustrate the manipulation of two different regions using a set of target anchor points, defined in red. 
Given that NFR does not employ any spatial shape constrains, optimizing anchor point positions also affects non-edited regions and the shape's identity as can be observed in \cref{fig:nfr} (NFR w/o reg.). To balance that we introduced a regularization that enforces the non-edited regions to remain unaffected (NFR w. reg.), which however, results in a manipulation performance drop as can be observed in the magnified areas. Similar to latent disentangled models, NFR struggles to locally edit expressions without impacting the unedited regions. It is also important to note that NFR, akin to SD-VAE and LED methods, necessitates an optimization scheme to achieve the target expression manipulations, which results in $20\times$ slower (3s vs 60s) performance compared to the proposed method. 
\section{Conclusion}
In this work we presented a diffusion 3D model for localized shape manipulation. The proposed method was trained using an inpainting inspired technique that guarantees local editing of the selected regions. Using this simple but intuitive approach our method outperforms current state-of-the-art disentangled manipulation methods and provides an effective solution to their limitations to ensure localized edits. Under a series of experiments, we show that the proposed method is able to manipulate facial and body parts as well as expressions controlled from a single or more anchor points. Beyond serving as an interactive 3D editing tool for digital artists, our method also offers notable applications in the field of aesthetic medicine.

\noindent\textbf{Acknowledgements.} S. Zafeiriou was supported by EPSRC Project DEFORM (EP/S010203/1) and GNOMON (EP/X011364). R.A. Potamias was supported by EPSRC Project GNOMON (EP/X011364).

%
%
\bibliographystyle{splncs04}
\bibliography{main}
\end{document}

%% file: Tables/main_results.tex
\begin{table*}[!ht]
\centering
\caption{Quantitative comparison between the proposed and the state-of-the-art methods on MimicMe \cite{papaioannou2022mimicme}, UHM \cite{ploumpis2019combining} and STAR \cite{STAR} datasets. All methods were trained on the same dataset for a fair comparison. All figures are measured in mm. }
\label{tab:results}
\resizebox{0.75\textwidth}{!}{ 
\begin{tabular}{l|ccc|ccc|ccc}
           & \multicolumn{3}{c|}{MimicMe \cite{papaioannou2022mimicme}  }            & \multicolumn{3}{c|}{UHM \cite{ploumpis2019combining}}                   & \multicolumn{3}{c}{STAR \cite{STAR}} \\
Method     & DIV $(\uparrow) $                     & FID $(\downarrow)$  & ID $(\downarrow)$    & DIV $(\uparrow)$                         & FID $(\downarrow)$   & ID  $(\downarrow)$   & DIV  $(\uparrow)$   & FID $(\downarrow)$   & ID $(\downarrow)$    \\ \hline \hline
M-VAE & 0.25                        & 1.21 & 0.09 & 0.61                         & 1.17 & 0.21 & 0.72    & 0.71  & 0.19  \\
SD \cite{foti2022sd_vae}   & 0.24                        & 7.81 & 0.84 & 0.53                         & 8.04 & 0.36 & 0.65   & 6.94  & 0.34  \\
LED \cite{foti2023led}    & 0.10                        & 3.39 & 0.23 & 0.43                         & 2.30 & 0.58 & 0.47   & 2.04  & 0.56  \\ \hline
\textbf{Proposed}       & \textbf{0.34} & \textbf{0.30} & \textbf{0.05} & \textbf{0.71} & \textbf{0.53} & \textbf{0.11} & \textbf{0.98} & \textbf{0.43} & \textbf{0.09} \\ \hline \hline
\end{tabular}}
\end{table*}